\documentclass[letter, 10 pt, conference]{ieeeconf}  
\usepackage{amsmath}
\usepackage{amsfonts}
\usepackage{color}
\usepackage{graphicx}
\usepackage[boxed]{algorithm2e}
\usepackage{mathrsfs}
\usepackage{amssymb}

\IEEEoverridecommandlockouts                              

\title{\LARGE \bf
Motion Planning with Safety Constraints and High-Level Task Specifications
}

\author{Seyedshams Feyzabadi \and Stefano Carpin
\thanks{S. Feyzabadi and S. Carpin are with the School of Engineering, University of California, Merced, CA, USA.\newline \newline
This work is supported by the National Institute of Standards and Technology under cooperative agreement 70NANB12H143.
Any opinions, findings, and conclusions or recommendations expressed in these materials are those of
the authors and should not be interpreted as representing the official policies, either expressly or
implied, of the funding agencies of the U.S. Government.
}
}

\begin{document}

\maketitle
\thispagestyle{empty}
\pagestyle{empty}

\section{Introduction}
The formalism of linear temporal logic (LTL) \cite{baier_2008} is increasingly being used to 
express task specifications in robotics, automation, and manufacturing.
Its expressiveness, coupled with its ease of use, makes it suitable for 
numerous scenarios. LTL alone, however, just expresses temporal relationships
and misses the ability to model the unavoidable uncertainty emerging in
 interactions with the physical world. To this end, Markov 
decision processes (MDPs) have been extensively used to formulate
solutions to a vast class of problems involving sequential stochastic
decision making under the hypothesis of state observability. 
In many practical situations, however, one is confronted
with multiple objective functions and MDPs alone are not suited in this scenario. Constrained Markov
Decision Processes (CMDPs)\cite{altman_1999}  offer a principled solution to this problem, whereby
one can determine policies optimizing one objective function while constraining
the costs associated with the remaining ones. Risk-aware motion planning has been tackled with CMDPs in \cite{feyzabadi_2014,feyzabadi_2015}.

In this paper we consider the case where both these formalisms are combined
together to determine control policies satisfying high level specifications
expressed in LTL while optimizing one or more functions as per the CMDP
framework. 

\section{Background}
\label{sec:back}
\subsection{Labeled CMDP}
A finite, labeled CMDP (LCMPD from now onwards) is an extension to CMDP (see \cite{altman_1999}) by adding $AP, L, F, \mathscr{S}$ variables to its original definition. Therefore, it is 
defined as $\mathcal{M}=(S, \beta, A, C_i, P, AP, L, F, \mathscr{S})$ where the extras to CMDPs are defined as

\begin{itemize}
\item $AP$ is a set of binary atomic propositions. 
\item $L$: $S \rightarrow 2^{AP}$ is a labeling function assigning to each state the set of atomic propositions true in the state.
\item $F\subset S$ is a (possibly empty) set of accepting states.
\item $\mathscr{S} \in S$ is a {\em  sink} state. 
An LCMDP may or may not have a sink state. In the latter case we will omit it when giving the definition.
\end{itemize}

\subsection{Co-safe LTL properties}
We consider a subset of LTL leading to so called co-safe LTL properties \cite{ding_2013}. Starting from a
set of atomic propositions $\Pi$, a co-safe LTL formula is built using the standard operators
and ($\wedge$), or ($\vee$), not ($\neg$) and the temporal operators {\em eventually} ($\Diamond$),
{\em next}  ($\bigcirc$), and {\em until} ($U$).
It is well known that given a co-safe LTL formula $\phi$, there exists a Deterministic Finite-state Automaton (DFA) accepting all and only 
the strings satisfying $\phi$ \cite{baier_2008}.

\section{Problem Formulation}
\label{sec:prel}

 Let   $\mathcal{M}=(S, \beta, A, C_i, P, AP, L, F)$ be an absorbing LCMDP with $n+1$ costs functions
$C_0,C_1,\dots,C_n$ and without any sink state,  and let $\phi$ be a syntactically co-safe LTL (sc-LTL) formula over $AP$.
Given  a probability $P_l$ and $n$ cost bounds $B_1,\dots,B_n$, determine a policy for $\mathcal{M}$ that:
\begin{itemize}
\item minimizes in expectation the $c_0(\pi,\beta)$;
\item for each cost $C_i$, ($1\leq i\leq n$), $c_i(\pi,\beta)\leq B_i$;
\item for every trajectory $\omega$, $\phi$  is satisfied with at least probability $P_l$.
\end{itemize}
An equivalent problem was studied in \cite{ding_2013}.  The solution we present in the following 
differs because we introduce a pruning step that effectively reduces the problem state space thus
leading to a much faster computation. Moreover, some of our definitions differ from \cite{ding_2013}
and lead to a more general solution.

\begin{figure*}
 \centering
 \includegraphics[width=0.225\textwidth]{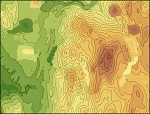}
 ~
 \includegraphics[width=0.22\textwidth]{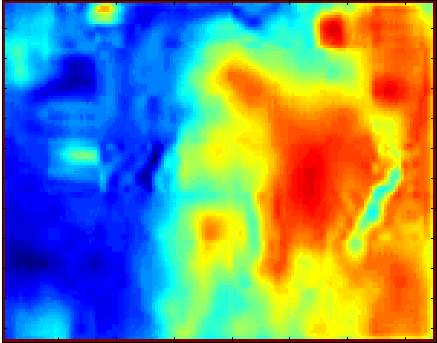}
 ~
 \includegraphics[width=0.22\textwidth]{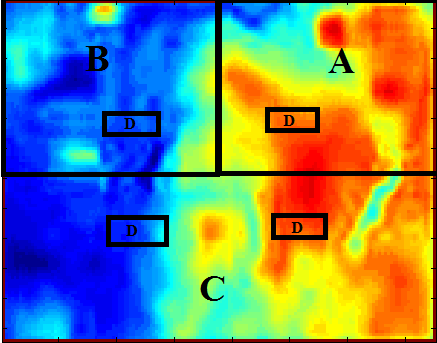}
 ~
 \includegraphics[width=0.22\textwidth]{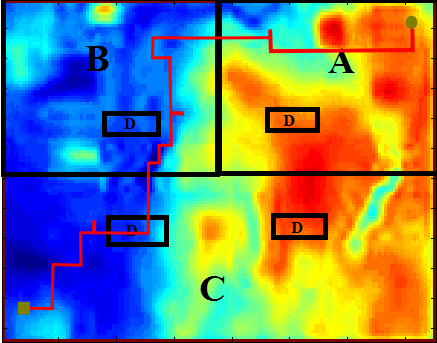}
 \caption{Experimental maps. Leftmost: A terrain map retrieved from web. Middle Left: The extracted risk map where cold colors represent regions with lower risks, 
 and warmer regions are riskier. Middle Right: regions with different labels. Rightmost: An example trajectory from start to goal.}
 \label{fig:maps}
\end{figure*}

\section{Proposed solution}
\label{sec:solution}
Our proposed solution contains three major steps that are as following: 
\begin{itemize}
 \item \textbf{Step 1}: A product between the given LCMDP and DFA associated with formula $\phi$ is calculated.  The product 
gives a new LCMDP for which a policy is computed. If the LCMDP contains $n_l$ states  and the DFA has $n_d$ states, 
the product LCMDP will consist of $n_l \cdot n_d $ states. 
  \item \textbf{Step 2}: In order to reduce the state space, a graph pruning algorithm is applied to the product LCMDP that removes some states and transitions 
  from the LCMDP while preserving the completeness of the solution. In other words in removes some parts of the graph that do not influence the final results. 
  A state may be used in the final policy if and only if there is a non-zero probability of 
  \begin{itemize}
   \item being reached from one of the initial states.
   \item and reaching the goal state.
   \item and reaching  or being reached from an accepting state. 
  \end{itemize}
  Otherwise, the state and its associated transitions can be removed from the LCMDP.

  \item \textbf{Step 3}:  The policy is obtained by solving a linear problem over the associated set of occupation measure variables.
\end{itemize}

\section{Experiments and Results}
\label{sec:exp}
To illustrate the method we propose, we consider an application of risk-aware motion planning. 
However, the current formulation introduces a high level task specification expressed with an LTL formula $\phi$.
We use a terrain map shown in the leftmost  picture of Figure \ref{fig:maps} and
a corresponding risk map was generated (see middle left picture). 

For every state four actions (up, down, left, right) are available, 
and each action succeeds with a certain probability influenced by the elevation difference
between neighboring cells. Risk is here defined as the probability of not succeeding when
executing an action. When an action does not succeed (i.e., when the desired motion does not occur),
the next position in the grid is chosen uniformly over the neighboring cells.
The map is divided into regions labeled as A,B,C, and D (see middle right panel in figure  \ref{fig:maps}).
The robot starts from a location in the top right and has to
reach an area in the bottom left corner of the map. The objectives are as follows:
\begin{itemize}
 \item cumulative total risk of the path has to be minimized;
 \item total path length has to be bounded by a constant $B=140$. To put this number into 
 perspective, the Manhattan distance between the start  and goal locations is 99.
 \item every path has to satisfy the formula $\phi = (A+B+C)^*D(D+C)^*$ with probability at least 0.7. 
\end{itemize}

The generated policy is used to extract multiple trajectories. Then we can assess
 how they match the mission objectives.
The rightmost panel in figure \ref{fig:maps} shows an example of path generated by the optimal policy.
The correctness of the formulation is confirmed. In 1000 trajectories generated with 
the policy returned by the linear program, the average risk is  486.1,
the average length is 122.5 and the formula $\phi$ is satisfied 703 times.


Finally, to evaluate the importance of the pruning algorithm we proposed,
we rescaled the same environment in order to generate equivalent problems
with a different number of variables in the linear program. Table \ref{table:pruning}
shows how the pruning step significantly reduces the time spent to solve the
linear program.
The first two columns show the number of variables in the linear program
with no pruning (first column -- NP) or with pruning (second column -- WP). The third and fourth column
show the time spent to solve the linear program with no pruning (third column) or with pruning (fourth column).

\begin{table}[htb]
\centering
\begin{tabular}{c|c|c|c}
\#Var NP & \#Var WP & Time (s) NP & Time (s) WP\\
\hline
1527                 &               887                               &                 54.16                 & 10.9\\
2035                 &               1322                             &                 87.43                & 21.18\\
2960                 &               2015                             &                 178.74            & 41.73\\
4182                 &               2935                             &                 372.57  & 83.21\\
\end{tabular}
\caption{Impact of pruning step. }
\label{table:pruning}
\end{table}

\section{Future Work}
In the future we will extend the problem by considering missions where multiple
task specifications can be included, each with different probability bounds.
In this case an iterated product between the LCMDP and multiple DFAs will be
necessary, thus exacerbating the formerly evidenced state-explosion problem.
In this situation, the value of the pruning algorithm we proposed will be even higher.

In general, by combining the formalism of constrained MDPs with linear temporal
logic it is possible to express multi objective planning problems that can be 
used to describe a rich set of automation and manufacturing tasks.

\bibliographystyle{plain}
\bibliography{references}

\begin{thebibliography}{1}

\bibitem{altman_1999}
E.~Altman.
\newblock {\em Constrained Markov Decision Processes}.
\newblock Stochastic modeling. Chapman \& Hall/CRC, 1999.

\bibitem{baier_2008}
C.~Baier and J.P Katoen.
\newblock {\em Principles of model checking}, volume~1.
\newblock MIT press Cambridge, 2008.

\bibitem{ding_2013}
X.~Ding, A.~Pinto, and A.~Surana.
\newblock Strategic planning under uncertainties via constrained markov
  decision processes.
\newblock In {\em Robotics and Automation (ICRA), 2013 IEEE International
  Conference on}, pages 4568--4575. IEEE, 2013.

\bibitem{feyzabadi_2014}
S.~Feyzabadi and S.~Carpin.
\newblock Risk-aware path planning using hierarchical constrained markov
  decision processes.
\newblock In {\em IEEE International Conference on Automation Science and
  Engineering}, 2014.

\bibitem{feyzabadi_2015}
S.~Feyzabadi and S.~Carpin.
\newblock Hcmdp: a hierarchical solution to constrained markov decision
  processes.
\newblock In {\em International Conference on Robotics and Automation (ICRA)},
  2015.

\end{thebibliography}

\end{document}